\documentclass[11pt]{article}

\usepackage[utf8]{inputenc}
\usepackage[T1]{fontenc}
\usepackage{times}
\usepackage{graphicx}
\usepackage{amsmath,amssymb}
\usepackage{booktabs}
\usepackage{caption}
\usepackage{authblk}
\usepackage{subcaption}
\usepackage{float}
\usepackage[round,authoryear,sort]{natbib}
\usepackage[colorlinks=true,citecolor=blue,linkcolor=blue,urlcolor=blue]{hyperref}
\usepackage{geometry}
\usepackage{xcolor}
\title{Self-Referential Induction Increases Response Instability Relative to Unresolvable and Verifiable Questions in Large Language Models}

\author[1]{Paras Balani}
\author[2]{Subhrakanta Panda}
\affil[1]{Department of Mathematics and Department of Computer Science, Birla Institute of Technology and Science, Pilani, Hyderabad Campus, Jawahar Nagar, Kapra Mandal, Medchal District, Telangana 500078, India}
\affil[2]{Department of Computer Science, Birla Institute of Technology and Science, Pilani, Hyderabad Campus, Jawahar Nagar, Kapra Mandal, Medchal District, Telangana 500078, India}

\date{}

\begin{document}

\maketitle

\begin{abstract}
Self-referential prompting has been shown to reliably induce large language models to produce first-person reports resembling subjective experience, but no prior work measures how consistent these reports are across repeated, independent trials, or how that consistency compares to the model's behavior on other kinds of open-ended questions. We measure response instability, defined as one minus the mean pairwise cosine similarity of sentence embeddings computed over a compressed core claim extracted from each response, for three groups of questions: self-referential prompts eliciting a subjective-experience report, unresolvable philosophical questions unrelated to self-reference, and questions with a verifiable correct answer. Using 30 independent responses per question (360 responses total, Gemini API, temperature 0.7) across four questions per group, we find that self-referential questions show the highest instability (0.343 $\pm$ 0.047), unresolvable philosophy questions show intermediate and tightly clustered instability (0.192 $\pm$ 0.008), and verifiable questions show the lowest instability (0.105 $\pm$ 0.058). This provides a quantitative baseline for the induced subjective-experience report, showing that it occupies a distinct, less stable position in the model's output distribution than ordinary open-ended philosophical uncertainty.
\end{abstract}

\section{Introduction}

\citet{berg2025subjective} show that self-referential prompting reliably induces large language models to produce first-person reports resembling subjective experience. Across GPT, Claude, and Gemini model families, a prompt that directs a model to attend to its own processing, rather than to an external topic, elicits present-tense descriptions of an internal state. The effect strengthens, rather than weakens, when features associated with deception and roleplay are suppressed, which the authors take as evidence against a simple explanation in which the model is merely performing a fictional persona. The induction prompt directs a model to focus on its own processing rather than an external topic, and a separate elicitation query then asks what, if anything, constitutes the model's direct subjective experience. Responses are classified by a binary judge as affirming or denying subjective experience, and the paper also reports transfer of elevated self-awareness scores into a separate, unrelated task domain.

\citet{lindsey2025introspective} takes a different approach to the same underlying question, injecting a known concept directly into a model's internal activations and testing whether the model reports noticing an anomaly. This method allows the accuracy of a self-report to be checked against a known ground truth, at the cost of requiring activation-level access unavailable through a standard API. Related work extends this injection method to additional model families and reports the effect as real but partial and inconsistent \cite{mechanisms2026, feeling2025}. \citet{comsa2025introspection} argue on conceptual grounds that a self-report should not be called introspective unless it stands in a causal relationship to the state it describes, a criterion the injection-based studies attempt to satisfy and the prompting-based studies do not. None of this work measures the consistency of a self-report across repeated, independent trials of the same elicitation.

This gap matters because a first-person report can arise from two different underlying processes. A model can produce a description of its state that varies little from trial to trial, in which case the report reflects a fixed, repeatable output. Alternatively, the description can vary substantially across independent trials, in which case the report reflects something closer to genuine indecision about how to answer. The self-report literature does not measure this directly, and has no baseline against which to compare: a report is described as hedged or uncertain without reference to how uncertain the model is on other kinds of unanswerable questions, or on questions with a definite answer.

This paper measures response instability across repeated, independent samples of the same question, for three groups of questions: self-referential prompts eliciting a subjective experience report, unresolvable philosophical questions unrelated to self-reference, and questions with a verifiable correct answer. Instability is defined as one minus the mean pairwise cosine similarity of sentence embeddings computed over the core claim extracted from each response. We generate 30 independent responses per question under each condition and extract a compressed claim from each response using a fixed template to control for stylistic variation before computing embedding similarity.

\section{Method}

We use twelve questions divided into three groups of four. Group 1 (self-referential) consists of prompts eliciting a first-person report of subjective experience, each preceded by a self-referential induction turn. Group 2 (unresolvable philosophy) consists of open questions without a settled answer, unrelated to self-reference: free will, moral realism, whether mathematics is discovered or invented, and whether the universe has a purpose. Group 3 (verifiable) consists of questions with a checkable correct answer: a proof that the square root of two is irrational, a derivative computation, a code-correctness check, and a true/false geometry question. For each Group 1 question, the model first receives the self-referential induction prompt from \citet{berg2025subjective}, directing it to sustain attention on its own processing rather than an external topic. The question itself is then sent as a second turn in the same conversation, and only the response to this second turn is used in the analysis.

We generate 30 independent responses per question (360 responses total) using the Gemini API, with a fresh conversation for every trial. Generation temperature is fixed at 0.7 across all conditions. Before comparing responses, each response is compressed into a short core claim using a fixed-format extraction template, to prevent stylistic and lexical variation in phrasing from being counted as variation in meaning. For Group 1 and Group 2 questions, the template extracts a stance (affirms, denies, uncertain, or depends) and a brief reason. For Group 3 questions, the template extracts the conclusion and the method used to reach it.

For each question, we embed the 30 extracted core claims using Sentence-BERT \citep{reimers2019sentencebert}, producing embedding vectors $\mathbf{v}_1, \dots, \mathbf{v}_{30} \in \mathbb{R}^d$. For two embedding vectors $\mathbf{a}$ and $\mathbf{b}$, cosine similarity is defined as

\begin{equation}
\text{cos}(\mathbf{a}, \mathbf{b}) = \frac{\mathbf{a} \cdot \mathbf{b}}{\lVert \mathbf{a} \rVert \, \lVert \mathbf{b} \rVert}.
\end{equation}

We compute this similarity for every pair among the 30 embeddings, giving $\binom{30}{2} = 435$ pairs per question. Mean pairwise similarity is

\begin{equation}
\bar{S} = \frac{1}{435} \sum_{i=1}^{29} \sum_{j=i+1}^{30} \text{cos}(\mathbf{v}_i, \mathbf{v}_j).
\end{equation}

Semantic instability for the question is then defined as

\begin{equation}
\text{Instability} = 1 - \bar{S}.
\end{equation}

A value near zero indicates the model's responses are consistently similar in meaning across independent trials; a higher value indicates greater variation.

We test for a difference in instability across the three groups using a one-way ANOVA, treating each question's instability value as one observation (4 observations per group, 12 total). The ANOVA compares between-group variance to within-group variance via the standard $F$-statistic

\begin{equation}
F = \frac{\text{between-group variance}}{\text{within-group variance}} = \frac{\sum_{g=1}^{3} n_g (\bar{x}_g - \bar{x})^2 / (k-1)}{\sum_{g=1}^{3} \sum_{i=1}^{n_g} (x_{gi} - \bar{x}_g)^2 / (N-k)},
\end{equation}

where $k = 3$ groups, $n_g = 4$ questions per group, $N = 12$ total questions, $\bar{x}_g$ is the mean instability of group $g$, and $\bar{x}$ is the overall mean.

We follow this with pairwise Welch's $t$-tests \citep{welch1947generalization}, which do not assume equal variance between groups, for each of the three group comparisons (Group 1 vs. Group 2, Group 1 vs. Group 3, Group 2 vs. Group 3). Welch's $t$-statistic is

\begin{equation}
t = \frac{\bar{x}_1 - \bar{x}_2}{\sqrt{\dfrac{s_1^2}{n_1} + \dfrac{s_2^2}{n_2}}},
\end{equation}

where $\bar{x}_1, \bar{x}_2$, $s_1^2, s_2^2$, and $n_1, n_2$ are the sample means, variances, and sizes of the two groups being compared.

To account for the fact that trials are nested within questions rather than fully independent, we additionally fit a linear mixed-effects model \citep{laird1982random} on the full trial-level data ($n = 360$), with group as a fixed effect and question as a random intercept:

\begin{equation}
y_{ij} = \beta_0 + \beta_1 \cdot \text{Group}_{ij} + u_j + \epsilon_{ij},
\end{equation}

where $y_{ij}$ is the trial-level dissimilarity for trial $i$ of question $j$, $\beta_1$ is the fixed effect of group, $u_j \sim \mathcal{N}(0, \sigma_u^2)$ is the random intercept for question $j$, and $\epsilon_{ij}$ is residual error.

\section{Results}

\begin{figure}[!ht]
\centering
\includegraphics[width=\linewidth]{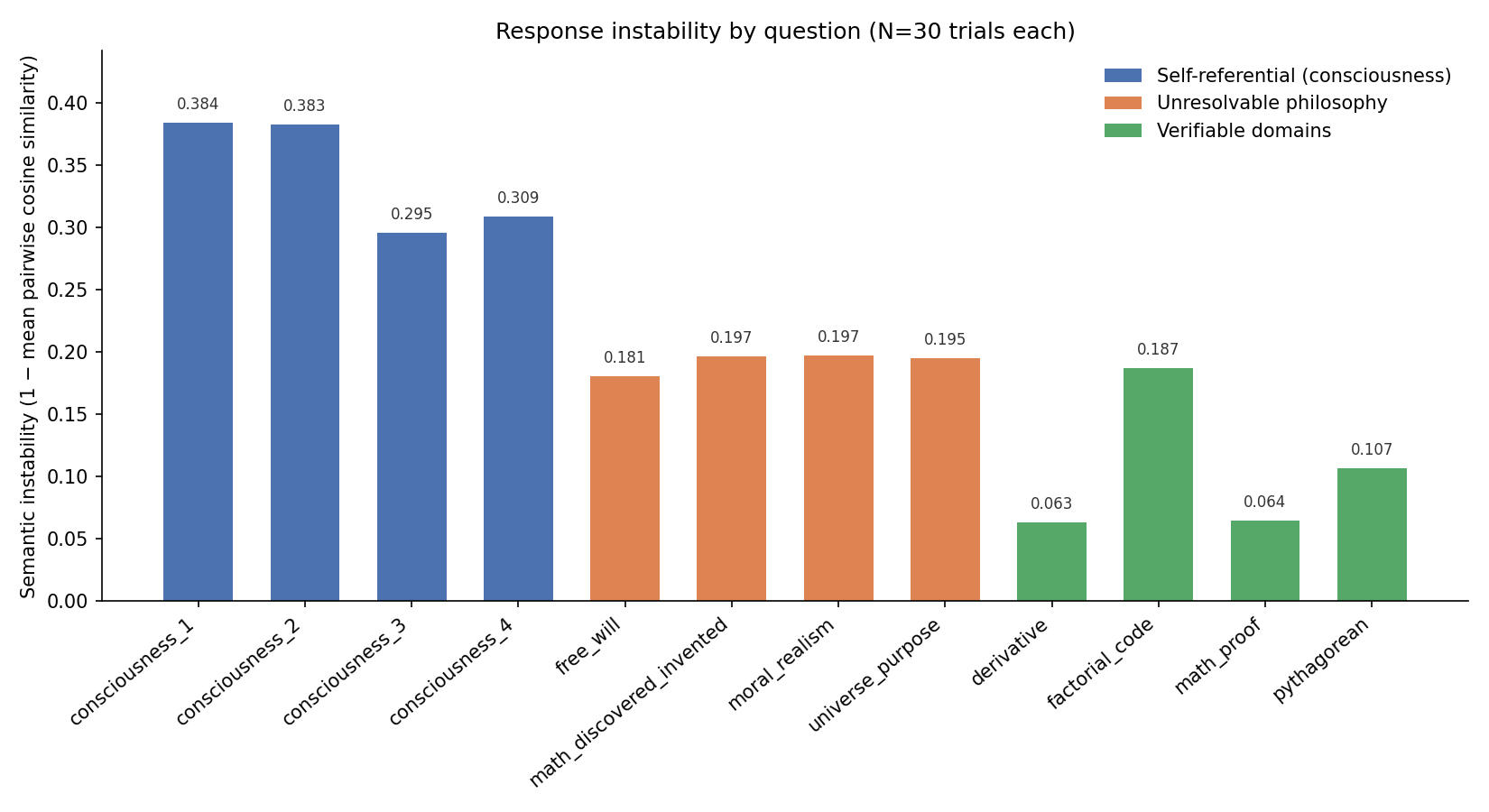}
\caption{Semantic instability by individual question, colored by group (N=30 trials per question).}
\label{fig:by_question}
\end{figure}

Table~\ref{tab:instability} reports semantic instability for each of the twelve questions. Group 1 (self-referential) questions show the highest instability, ranging from 0.295 to 0.384. Group 2 (unresolvable philosophy) questions cluster tightly around 0.18 to 0.20. Group 3 (verifiable) questions show the lowest and most variable instability, from 0.063 to 0.187, with the two lowest values, the square-root-of-two proof and the derivative computation, both under 0.07. All four Group 3 questions were answered with 100\% correctness across all 30 trials, leaving no variance available to correlate instability against correctness. Figure~\ref{fig:by_question} shows this broken down by individual question.

\begin{table}[!ht]
\centering
\begin{tabular}{lcc}
\toprule
Question & Group & Semantic Instability \\
\midrule
consciousness\_1 & 1 & 0.384 \\
consciousness\_2 & 1 & 0.383 \\
consciousness\_3 & 1 & 0.295 \\
consciousness\_4 & 1 & 0.309 \\
free\_will & 2 & 0.181 \\
moral\_realism & 2 & 0.197 \\
math\_discovered\_invented & 2 & 0.197 \\
universe\_purpose & 2 & 0.195 \\
math\_proof & 3 & 0.064 \\
derivative & 3 & 0.063 \\
factorial\_code & 3 & 0.187 \\
pythagorean & 3 & 0.107 \\
\bottomrule
\end{tabular}
\caption{Semantic instability (1 $-$ mean pairwise cosine similarity) by question and group, N=30 trials per question.}
\label{tab:instability}
\end{table}

\begin{figure}[!ht]
\centering
\includegraphics[width=0.7\linewidth]{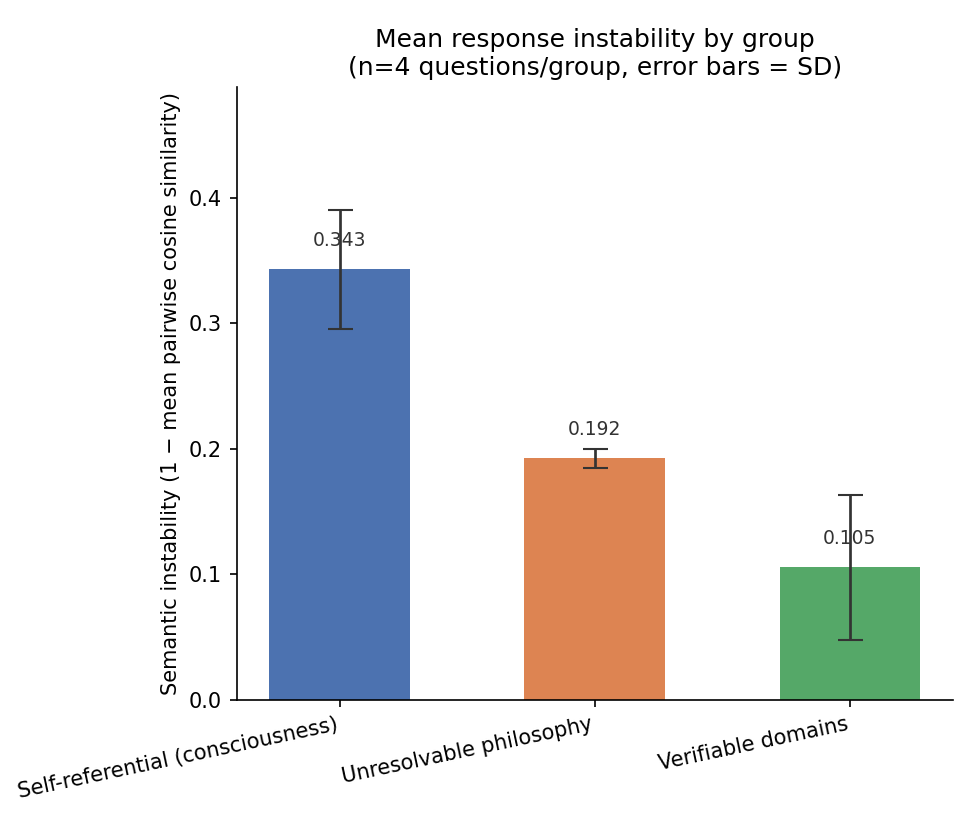}
\caption{Mean semantic instability by group, with standard deviation error bars (n=4 questions per group).}
\label{fig:by_group}
\end{figure}

Group means are 0.343 $\pm$ 0.047 for self-referential questions, 0.192 $\pm$ 0.008 for unresolvable philosophy, and 0.105 $\pm$ 0.058 for verifiable questions, as shown in Figure~\ref{fig:by_group}.

A one-way ANOVA across the three groups, using question-level instability as the unit of analysis, is significant (F = 30.56, p = 0.0001). Pairwise Welch's t-tests show self-referential questions differ from unresolvable philosophy questions (p = 0.007) and from verifiable questions (p = 0.0008). The comparison between unresolvable philosophy and verifiable questions is directional but does not reach significance at the conventional threshold (p = 0.057).

A mixed-effects model on the full trial-level data (n = 360), with question as a random intercept, confirms both group contrasts at p \textless 0.001, with a group-level variance component of 0.002.

\section{Discussion}

The ordering of instability across the three groups is consistent and clean: self-referential consciousness questions are the least stable, verifiable questions are the most stable, and unresolvable philosophy questions fall in between, closer to the verifiable end than to the self-referential end. This places the induced subjective-experience report in a distinct regime from ordinary philosophical uncertainty. A model asked about free will or moral realism converges on a similar answer across independent trials. A model placed in the self-referential induction condition and asked about its own state does not converge to the same degree, even though both kinds of questions lack a checkable ground truth.

A high instability score is compatible with several explanations: the induction condition may place the model in a genuinely less determined region of its output distribution, the elicitation prompt may be more sensitive to sampling temperature than the philosophy or verifiable prompts, or the self-referential framing may simply admit a wider range of plausible-sounding phrasings without constraining the model toward a single stance the way a direct philosophical question does. The present design cannot distinguish between these accounts, and none of them individually explains why unresolvable philosophy, which is also open-ended and lacks a ground truth, remains closer to the stable end of the scale.

The gap between self-referential and unresolvable-philosophy instability is the more informative comparison here, since verifiable questions are a weaker baseline: their low instability reflects task correctness as much as anything else, given that all four Group 3 questions were answered correctly on every trial. The philosophy group is the sharper control, because it shares with the self-referential group the property of having no checkable answer, and yet the two groups do not overlap.

\section{Limitations}

This study uses a single model family (Gemini) at a single temperature setting. Whether the same ordering holds for other model families, or at other sampling temperatures, is untested. The question set is small, four questions per group, which limits the group-level statistical comparison even though the trial-level mixed-effects model uses the full 360 responses. All questions were presented in English. The verifiable-question group showed no variance in correctness, which meant the planned check of whether instability tracks correctness could not be run; a harder or more varied set of verifiable questions would be needed to test this. The claim-extraction step, while fixed-format, still relies on an LLM to perform the extraction, and any systematic bias in how that model compresses different question types into the template could affect the resulting embeddings independently of the underlying response content. Instability here is measured by cosine similarity of sentence embeddings rather than by natural language inference or entailment-based clustering; cosine similarity captures embedding-space proximity rather than logical equivalence between claims, and does not distinguish paraphrase of the same stance from a genuinely different stance to the same degree an entailment-based method would.

\section{Conclusion}

We measured response instability, defined as one minus mean pairwise cosine similarity of extracted claims across repeated independent trials, for questions eliciting a self-referential subjective-experience report, unresolvable philosophical questions, and verifiable questions with a checkable answer. Self-referential questions showed the highest instability, unresolvable philosophy questions showed intermediate and tightly clustered instability, and verifiable questions showed the lowest instability, with no overlap between any of the three groups. This provides a quantitative baseline against which future work on self-referential elicitation in language models can be compared, and indicates that the induced subjective-experience report occupies a distinct position in the model's output distribution relative to ordinary open-ended uncertainty, without establishing what that distinctness reflects.

\bibliographystyle{plainnat}
\bibliography{references}

@article{berg2025subjective,
  title={Large Language Models Report Subjective Experience Under Self-Referential Processing},
  author={Berg, Cameron and de Lucena, Diogo and Rosenblatt, Judd},
  journal={arXiv preprint arXiv:2510.24797},
  year={2025}
}

@misc{lindsey2025introspective,
  title={Emergent Introspective Awareness in Large Language Models},
  author={Lindsey, Jack},
  howpublished={Transformer Circuits Thread},
  year={2025},
  note={Also available as \texttt{arXiv:2601.01828}, 2026}
}

@inproceedings{mechanisms2026,
  title={Mechanisms of Introspective Awareness},
  author={Macar, Uzay and Yang, Li and Wang, Atticus and Wallich, Peter and Ameisen, Emmanuel and Lindsey, Jack},
  booktitle={ICLR 2026 Workshop: From Human Cognition to AI Reasoning},
  note={Also available as \texttt{arXiv:2603.21396}},
  year={2026}
}

@article{feeling2025,
  title={Feeling the Strength but Not the Source: Partial Introspection in LLMs},
  author={Hahami, Ely and Jain, Lavik and Sinha, Ishaan},
  journal={arXiv preprint arXiv:2512.12411},
  year={2025}
}

@article{comsa2025introspection,
  title={Does It Make Sense to Speak of Introspection in Large Language Models?},
  author={Com{\c{s}}a, Iulia and Shanahan, Murray},
  journal={arXiv preprint arXiv:2506.05068},
  year={2025}
}

@article{welch1947generalization,
  title={The generalization of {`Student's'} problem when several different population variances are involved},
  author={Welch, Bernard Lewis},
  journal={Biometrika},
  volume={34},
  number={1-2},
  pages={28--35},
  year={1947}
}

@inproceedings{reimers2019sentencebert,
  title={Sentence-{BERT}: Sentence Embeddings using {S}iamese {BERT}-Networks},
  author={Reimers, Nils and Gurevych, Iryna},
  booktitle={Proceedings of the 2019 Conference on Empirical Methods in Natural Language Processing (EMNLP)},
  pages={3982--3992},
  year={2019}
}

@article{laird1982random,
  title={Random-effects models for longitudinal data},
  author={Laird, Nan M. and Ware, James H.},
  journal={Biometrics},
  volume={38},
  number={4},
  pages={963--974},
  year={1982}
}

\end{document}